\documentclass[letterpaper, 10 pt, conference]{ieeeconf}  % Comment this line out if you need a4paper

\IEEEoverridecommandlockouts                              % This command is only needed if 
\usepackage{amsmath,amsfonts}
\usepackage{algorithmic}
\usepackage{array}
\usepackage[caption=false,font=normalsize,labelfont=sf,textfont=sf]{subfig}
\usepackage{textcomp}
\usepackage{stfloats}
\usepackage{url}
\usepackage{verbatim}
\usepackage{graphicx}
\usepackage{hyperref}
\usepackage{caption}
\usepackage{graphicx}
\usepackage{float} 
\usepackage{subcaption}
\usepackage{booktabs}
\newsavebox\CBox
\usepackage{graphicx} 
\usepackage{xcolor}
\usepackage{xpatch}
\usepackage{multirow}
\usepackage {hyperref}
\title{\LARGE \bf
Distribution-aware Noisy-label Crack Segmentation
}

\author{Xiaoyan Jiang$^{1*}$, Xinlong Wan$^{1*}$, Kaiying Zhu$^{2}$, Xihe Qiu$^{1}$, and Zhijun Fang$^{3\dagger}$% <-this % stops a space
\thanks{* Equal contribution.}% <-this % stops a space
\thanks{$\dagger$ Corresponding author: {\tt\small zjfang@dhu.edu.cn}}
\thanks{$^{1}$School of Electronic and Electrical Engineering, Shanghai University of Engineering Science,  {{\tt\small xiaoyan.jiang@sues.edu.cn}}.
$^{2}$SenseTime. $^{3}$School of Computer Science and Technology, Donghua University.}%
}

\begin{document}

\maketitle
\thispagestyle{empty}
\pagestyle{empty}

%%%%%%%%%%%%%%%%%%%%%%%%%%%%%%%%%%%%%%%%%%%%%%%%%%%%%%%%%%%%%%%%%%%%%%%%%%%%%%%%
\begin{abstract}
Road crack segmentation is critical for robotic systems tasked with the inspection, maintenance, and monitoring of road infrastructures. 
Existing deep learning-based methods for crack segmentation are typically trained on specific datasets, which can lead to significant performance degradation when applied to unseen real-world scenarios. 
To address this, we introduce the SAM-Adapter, which incorporates the general knowledge of the Segment Anything Model (SAM) into crack segmentation, demonstrating enhanced performance and generalization capabilities. 
However, the effectiveness of the SAM-Adapter is constrained by noisy labels within small-scale training sets, including omissions and mislabeling of cracks. 
In this paper, we present an innovative joint learning framework that utilizes distribution-aware domain-specific semantic knowledge to guide the discriminative learning process of the SAM-Adapter. 
To our knowledge, this is the first approach that effectively minimizes the adverse effects of noisy labels on the supervised learning of the SAM-Adapter.
Our experimental results on two public pavement crack segmentation datasets confirm that our method significantly outperforms existing state-of-the-art techniques. Furthermore, evaluations on the completely unseen CFD dataset demonstrate the high cross-domain generalization capability of our model, underscoring its potential for practical applications in crack segmentation.
Code is available at \href{https://github.com/sky-visionX/CrackSegmentation}{https://github.com/sky-visionX/CrackSegmentation}.
% Road crack segmentation is crucial for robotic systems targeting inspections, maintenance, and monitoring of structures of roads.
% Existing state-of-the-art deep learning-based crack segmentation methods normally trained on specific dataset, leading to a severe performance decline in unseen practical scenarios. 
% We adopt SAM-Adapter to inject general knowledge of Segment Anything Model (SAM) to crack segmentation, showing a promising performance and generalization ability.
% However, noisy labels in the small-scale training set, that is, omission and mislabel of cracks, limiting the power of adapting SAM to crack segmentation.
% In this paper, we present an efficient joint learning framework for crack segmentation by distribution-aware domain-specific semantic knowledge guiding the discriminate learning of SAM-Adapter.
% To the best of our knowledge, we are the first to propose solutions to efficiently mitigate the negative impact of noisy crack labels on the supervised SAM-Adapter learning model.
% Experimental results on the pavement crack segmentation datasets show that our approach surpasses the-state-of-art methods by a large margin. 
% In addition, the proposed model shows high cross-domain generalization ability for potential practical crack segmentation.
% Code is available.
\end{abstract}

%%%%%%%%%%%%%%%%%%%%%%%%%%%%%%%%%%%%%%%%%%%%%%%%%%%%%%%%%%%%%%%%%%%%%%%%%%%%%%%%
\section{INTRODUCTION}
{T}{o} avert serious road damage and catastrophic driving accidents, it is imperative to carry out regular inspections and timely maintenance of all kinds of road surfaces \cite{ref1}.
Various types of robots in the world are equipped with multiple cameras, which observe directly the road to capture images \cite{ref2}. 
The target is to provide an automatic and intelligent road health assessment system to replace the manual road inspection.
Among pavement defects, cracks are the most common but challenging type to be detected due to the intra-class difference and the texture-less characteristic \cite{ref3}. 
In contrast to crack detection, pavement crack segmentation provides pixel-level classification, offering the exact location and shape of the cracks \cite{ref4}. 
Current mainstream state-of-the-art crack segmentation algorithms are based on deep learning model architectures, such as, Convolutional Neural Network (CNN) and Transformer \cite{ref5,ref6,ref7}. 
However, public datasets are rare and small scale, which makes current crack segmentation trained on specific dataset has a far-from-practice performance and a poor generalization ability.

\begin{figure}[!t]
	\centering
	\begin{minipage}{0.32\linewidth}
		\centering
		\includegraphics[width=2.8cm,height=1.6cm]{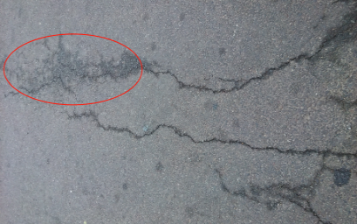}
	\end{minipage}
	\begin{minipage}{0.32\linewidth}
		\centering
		\includegraphics[width=2.8cm,height=1.6cm]{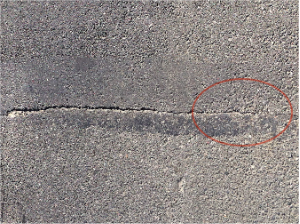}
	\end{minipage}
	\begin{minipage}{0.32\linewidth}
		\centering
		\includegraphics[width=2.8cm,height=1.6cm]{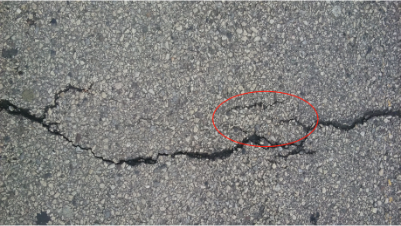}
	\end{minipage}
 % \vspace{1.5mm}
    % \caption{mother of chutian[XJ5]}
    % \label{da_chutian}
     \vspace{1.5mm}
     {\fontsize{9pt}{1pt}\selectfont (a) Original images}
	% \vspace{1mm}
	%\qquad
 
	\begin{minipage}{0.32\linewidth}
		\centering
		\includegraphics[width=2.8cm,height=1.6cm]{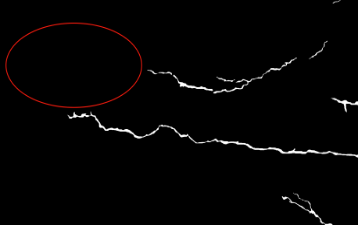}
	\end{minipage}
	\begin{minipage}{0.32\linewidth}
		\centering
		\includegraphics[width=2.8cm,height=1.6cm]{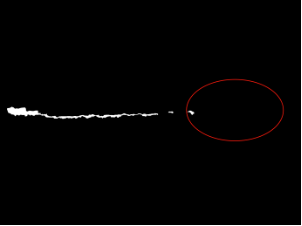}
	\end{minipage}
        \begin{minipage}{0.32\linewidth}
		\centering
		\includegraphics[width=2.8cm,height=1.6cm]{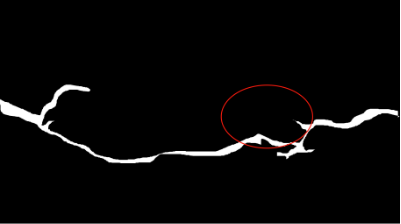}
	\end{minipage}
         % \vspace{1.25mm}
 	% \vspace{1.5mm}
        {\fontsize{9pt}{1pt}\selectfont (b) Annotated labels}
    \caption{Examples of noisy labeled samples in the Crack500 \cite{ref8} dataset. 
    %Image in the second row are the corresponding labels of the first row.
    }
    \label{Fig1}
\end{figure}

Importantly, we found that cracks labeling is a tedious and challenging task for human annotators with subjective bias. 
Accurately identifying thin cracks isn't always straightforward due to various factors in image acquisition, such as resolution, sharpness, motion blur, and illumination conditions. 
As a result, annotators always cannot correctly identifying cracks, leading to annotation errors, such as, missing labelling, and false positive labels. 
Take the largest publicly available annotated crack segmentation dataset as an example, within the 250 images for training \cite{ref8}, we can identify under-annotation of thin cracks in multiple samples, as depicted in Fig \ref{Fig1}.
Recently, Xu et al. \cite{ref8} explore the affect of different types of annotation errors on the performance of crack segmentation by simulating errors. They argue that the data under-annotation has the most substantial negative affect on the performance of supervised models. 

To the best of our knowledge, however, we have not identified any work deliberating on how to handle with the negative impact of noisy labels on the crack segmentation model training.
%To avoid the high cost of pixel-level annotation, previous works have attempted to address road crack segmentation using unsupervised methods \cite{ref10}. 
%Due to challenges, such as image acquisition resolution, lighting variations, intra-class variance of cracks, and background noise, robust feature extraction becomes exceedingly difficult.  
%In contrast, supervised methods are more effective in addressing the diversity and variability of road surface crack images, enhancing the consistency and reliability of crack segmentation.  
At present, majority researches focus on the supervised crack segmentation enhancement on the single dataset, such as Crack500. 
Models trained on a small, sole dataset is prone to overfitting, with side-effect caused by mis-annotated data.
%and lacks general knowledge of image samples. 
However, rare works consider the generalization ability of crack recognition across scenarios, which is practically important. 
%All these significantly restrict the practical application of deep learning in the field of road crack segmentation. 

Nowadays, SAM \cite{ref9} is a groundbreaking large-scale pre-training segmentation model in the visual field. 
% Training on extensive image data makes SAM absorbs general foundation knowledge, demonstrating good segmentation performance and generalization across diverse tasks.  
% But, existing research has proved that directly using SAM in downstream scenarios often fails to achieve the expected outcomes \cite{ref13}.
% SAM-Adapter integrates the general knowledge of SAM with downstream domain-specific information to obtain superior results on multiple tasks \cite{ref13}.
While it demonstrates good segmentation performance and generalization across diverse tasks after training on extensive image data, using SAM directly in downstream scenarios often fails to achieve expected outcomes. To address this issue, SAM-Adapter integrates the general knowledge of SAM with domain-specific information to achieve superior results on multiple tasks \cite{ref10}. 

\begin{figure}[!t]
	\centering
 	\begin{minipage}{0.2\linewidth}
		\centering
		\includegraphics[width=1.8cm,height=1.6cm]{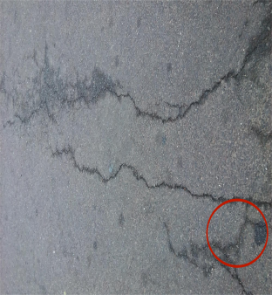}
	\end{minipage}
	\begin{minipage}{0.2\linewidth}
		\centering
		\includegraphics[width=1.8cm,height=1.6cm]{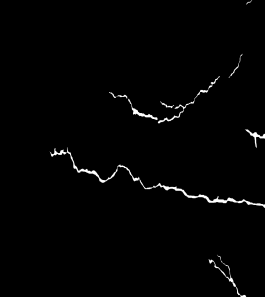}
	\end{minipage}
	\begin{minipage}{0.2\linewidth}
		\centering
		\includegraphics[width=1.8cm,height=1.6cm]{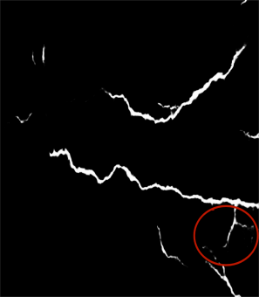}
	\end{minipage}
	\begin{minipage}{0.2\linewidth}
		\centering
		\includegraphics[width=1.8cm,height=1.6cm]{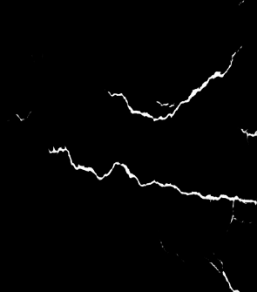}
	\end{minipage}
        \vspace{1mm}

      % {\fontsize{9pt}{1pt}\selectfont (a) Original Image}

	% 
	% 
	\begin{minipage}{0.2\linewidth}
		\centering
		\includegraphics[width=1.8cm,height=1.6cm]{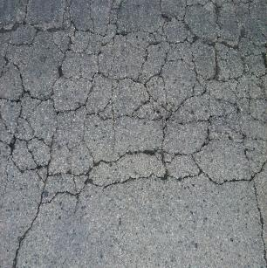}
		\centerline{\fontsize{7.5pt}{\baselineskip}\selectfont Original images} %修改字体大小
	\end{minipage}
	\begin{minipage}{0.2\linewidth}
		\centering
		\includegraphics[width=1.8cm,height=1.6cm]{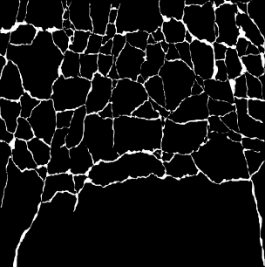}
  		\centerline{\fontsize{7.5pt}{\baselineskip}\selectfont Annotated labels}
	\end{minipage}
        \begin{minipage}{0.2\linewidth}
		\centering
		\includegraphics[width=1.8cm,height=1.6cm]{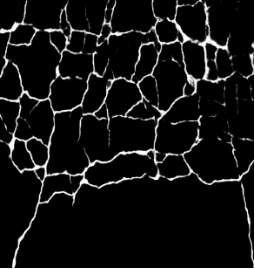}
            \centerline{\fontsize{7.5pt}{\baselineskip}\selectfont Epoch=25}
	\end{minipage}
        \begin{minipage}{0.2\linewidth}
		\centering
		\includegraphics[width=1.8cm,height=1.6cm]{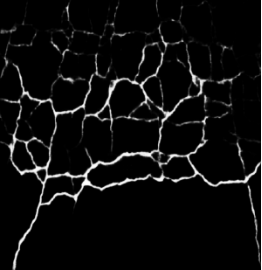}
            \centerline{\fontsize{7.5pt}{\baselineskip}\selectfont Epoch=50}
	\end{minipage}
         \vspace{1.5mm}
 	% \vspace{1.5mm}
    \caption{Upper row: mislabeled sample in SAM-Adapter training. Mislabeled pixels disappear as training goes on. Lower row: changes of test sample predictions corresponding to training epoches.}
    \label{Fig2}
\end{figure}

Is SAM's general knowledge helpful for crack segmentation and the model's generalization ability?
%domain knowledge through adaptation to address the issue of insufficient crack feature recognition ability and poor generalization ability in cross-domain? 
Since noisy labels occupy a large portion in the small-scale crack dataset labels, can the SAM-Adapter model suppresses the negative impact? 
In this paper, we introduce and analyse SAM-Adapter in pavement crack segmentation task.
Experiments demonstrate that SAM-Adapter can significantly enhance the crack recognition ability and generalization ability in cross-domain. 
However, as shown in Fig \ref{Fig2}, with the increase of training epoches, numerous small cracks gradually fade into the background. Due to the interference of noise labels, the SAM-Adapter model  learns incorrect features and mistakenly identifies some cracks as the background, leading to an overfit on noisy labels.

\textit{How to enhance the SAM-Adapter model's capacity of cross-domain crack recognition while mitigating the adverse effect of noisy labels on the model? }
%To this end, we contemplated more profoundly. 
For each individual image, we observe that the characteristics of diverse cracks indeed are visually distinct from the background. 
Theoretically, the pixel-level semantic representations of different cracks extracted by SAM-Adapter should possess high correlation and differ from the background features. 
However, cracks of different image samples cannot be uniformly described from appearance, which lead to the low performance of current state-of-the-art discriminative learning-based crack segmentation methods.
Observing the evolution process of training prediction shown in Fig \ref{Fig2}, accurate pixel-level semantic representations can be exploited in suitable epoches and used to guide the learning of SAM adaptation module. 
%During the training process, domain-specific semantic information generated by the unpolluted pixel-level semantic representations are coupled with SAM-Adapter discriminate model in a self-supervised manner to alleviate the negative impact of noise labels. 
%Therefore, we envision whether the pixel-level semantic representations of pavement crack samples learned by the model can be exploited to mine the similarity relationship among their representations for semantic modelling. 

In this paper, we present a novel crack segmentation training framework coupling domain-specific semantic representation extraction with discriminative learning of the SAM-Adapter model.
To protect the individual characteristics of cracks in each image,  
we infer the category semantic representation of every image as Mixture of Gaussian distributions (MoG).
Compared with works using Gaussian Mixture Model for all samples in the training dataset \cite{ref11}, our strategy is more memory efficient and accurate.
The parameters of the MoG, that is, means and covariances of the Gaussian distributions, are estimated using the Expectation-Maximization (EM) algorithm in an unsupervised manner.
Notably, MoG-based representation learning module can produce uncontaminationed labels, which %jointly supervise the training process of the SAM-Adapter model given the manual annotated labels.
can adaptively optimize the SAM-Adapter model's learning direction and mitigate the negative impact of noisy labels on the model.
Experimental results show that our approach significantly outperforms the state-of-the-art methods based on either CNN, Transformer, or large model adaptation on the commonly used crack segmentation datasets. 

The contributions are summarized as follows:
(1) We introduce SAM-Adapter as an adaptive scheme for road crack segmentation, integrating the domain information of cracks with the general knowledge of SAM. We assess its generalization ability and analyze its crack segmentation performance under noisy labels.
(2) To the best of our knowledge, we are the first who explicitly suppress the negative affect of noisy labels on crack segmentation. The domain-specific category semantic label is learned by an end-to-end probabilistic model-based on MoG, which can both supervise and are adaptively influenced by the discriminative SAM-Adapter learning.
(3) We propose a novel framework that models the semantic representation of each image as a mixture of Gaussian distributions, reducing the impact of noisy labels. Results show that our method outperforms state-of-the-art approaches on two public datasets.
% Evaluation results on two standard datasets show that our approach surpasses the state-of-the-art methods by large margins. 
% The proposed approach can effectively mitigate the negative affect of noise labels. 
% Moreover, without training on the CFD dataset, our method still achieves super state-of-the-art performance, which indicates the strong generalization ability of our model across scenarios. 

\section{RELATED WORKS}
\subsection{Crack segmentation}
% The mainstream crack segmentation methods adopt deep learning-based semantic segmentation framework to achieve precise and fine-grained pixel-level classification of road cracks \cite{ref14,ref15,ref16}. A considerable number of pavement crack segmentation works \cite{ref20,ref22} adopt the network architectures, such as, fully convolutional neural network (FCN), U-Net, and SegNet \cite{ref19}. 
The mainstream crack segmentation methods use deep learning-based semantic segmentation framework for precise and fine-grained pixel-level classification of road cracks \cite{ref12,ref13,ref14}. Many pavement crack segmentation works \cite{ref15,ref16} utilize network architectures, such as, fully convolutional neural network (FCN), U-Net, and SegNet \cite{ref17}.
Due to the excellent global features extraction and context information capturing ability, Transformers \cite{ref18,ref19} are  adopted for crack segmentation to achieve more refined outcomes \cite{ref20,ref21,ref22}.
% Segformer, utilizing multi-head self-attention and hierarchical features, excels in capturing image details, significantly boosting crack segmentation over traditional CNNs.\cite{ref27,ref28}. 
% Although cracks are texture-less parts in the images and easily affected by the complicate background clutters, the encoder of CrackFormer-II \cite{ref31} obtains more potent feature expression capacity.
% Meanwhile, its decoder achieves high-precision segmentation of cracks by fusing multi-scale features, being the most state-of-the-art approach in this domain.
Segformer \cite{ref21}, with multi-head self-attention and hierarchical features, excels in capturing image details and significantly improves crack segmentation compared to CNNs. 
The encoder of CrackFormer-II \cite{ref23} has a more powerful feature expression capacity by fusing multi-scale features, being the most state-of-the-art approach in this domain.

However, the current crack segmentation works mostly focus on performance improvement on a certain dataset, that is, enhancing the recognition ability of cracks on the specific dataset.
In practice, the trained model has a severe performance decay when changing to another unseen scenario.
But, studies on the model's generalization ability for crack recognition in cross-domain are relatively rare. 
%Additionally, since direct adopting SAM-Adapter

\subsection{Learning with noisy labels}
Rare defect segmentation studies explored the effect of noisy labels on the performance of deep learning-based models.
Recently, the influence of noisy labels on the performance of crack segmentation was investigated by simulating different types of annotation errors \cite{ref24}. 
They found that under-annotation has the largest negative effects on the model's performance. However, no research studies how to suppress negative effects of noisy labels on dense crack segmentation. 
% Most of the existing research on noisy labels focuses on the image classification task. 
% Currently, studies of learning from noisy labels mainly have the following three strategies: correct noisy labels to improve the learning performance \cite{ref32,ref33,ref33}, weaken the influence of noisy samples on model training through methods of re-weighting samples or sample selection \cite{ref34,ref35}, and improve the learning performance by designing robust loss functions \cite{ref36,ref37,ref38}.
Most existing research on noisy labels has been in the context of image classification tasks. Strategies for learning from noisy labels include correcting noisy labels \cite{ref25,ref26}, reducing their influence through re-weighting or sample selection \cite{ref27,ref28}, and designing robust loss functions to improve learning performance \cite{ref29,ref30,ref31}.
 
In this paper, we explicitly consider the generalization ability of the model across scenarios and the problem of noisy labels in the training data of crack samples.
A through analysis for applying SAM-Adapter to the texture-less crack segmentation task with noisy labels are presented.

\subsection{Mixture of Gaussians}
Gaussian distribution is a classic probability distribution function to mimic natural phenomenons.
% The distribution of data is represented by two important parameters: the mean, describing the data central position, and the variance, describing the degree of data dispersion.
% While the Mixture of Gaussians (MoG) fits the complex structure of data by using the combination of multiple Gaussian distributions \cite{ref39,ref40}. 
% Each Gaussian distribution can specifically capture a local structure in the data, endowing it with the ability to model the data distribution of any shape \cite{ref41,ref42}. 
% Recently, MoG has been attempted to be combined with deep learning techniques in the field of computer vision for modeling the latent structure distribution of deep features \cite{ref43,ref44,ref45}.
The Mixture of Gaussians fits complex data structures by combining multiple Gaussian distributions \cite{ref32,ref33}. Each Gaussian distribution captures a local structure in the data, allowing it to model data distributions of any shape \cite{ref34,ref35}. 
Recently, MoG has been combined with deep learning techniques in computer vision to model the latent structure distribution of deep features \cite{ref36,ref37,ref38}.

The intra-class variance of crack types makes the segmenting of cracks from complicated pavements background extremely challenging. 
Moreover, due to the mislabeled samples and scarcity of public crack datasets, models are easily corrupted because of the side-effect \cite{ref36}. 
We observed that SAM Adapter captures good features at the beginning of the training, but, noisy labels give negative effect as training going on.
In this paper, MoG, as the generative model, is combined to guide the training, leading to a more promising performance.

\section{The PROPOSED METHOD}
\begin{figure}[!t]
	\centering
 	\begin{minipage}{0.9\linewidth}
		\centering
		\includegraphics[width=8.5cm,height=4.2cm]{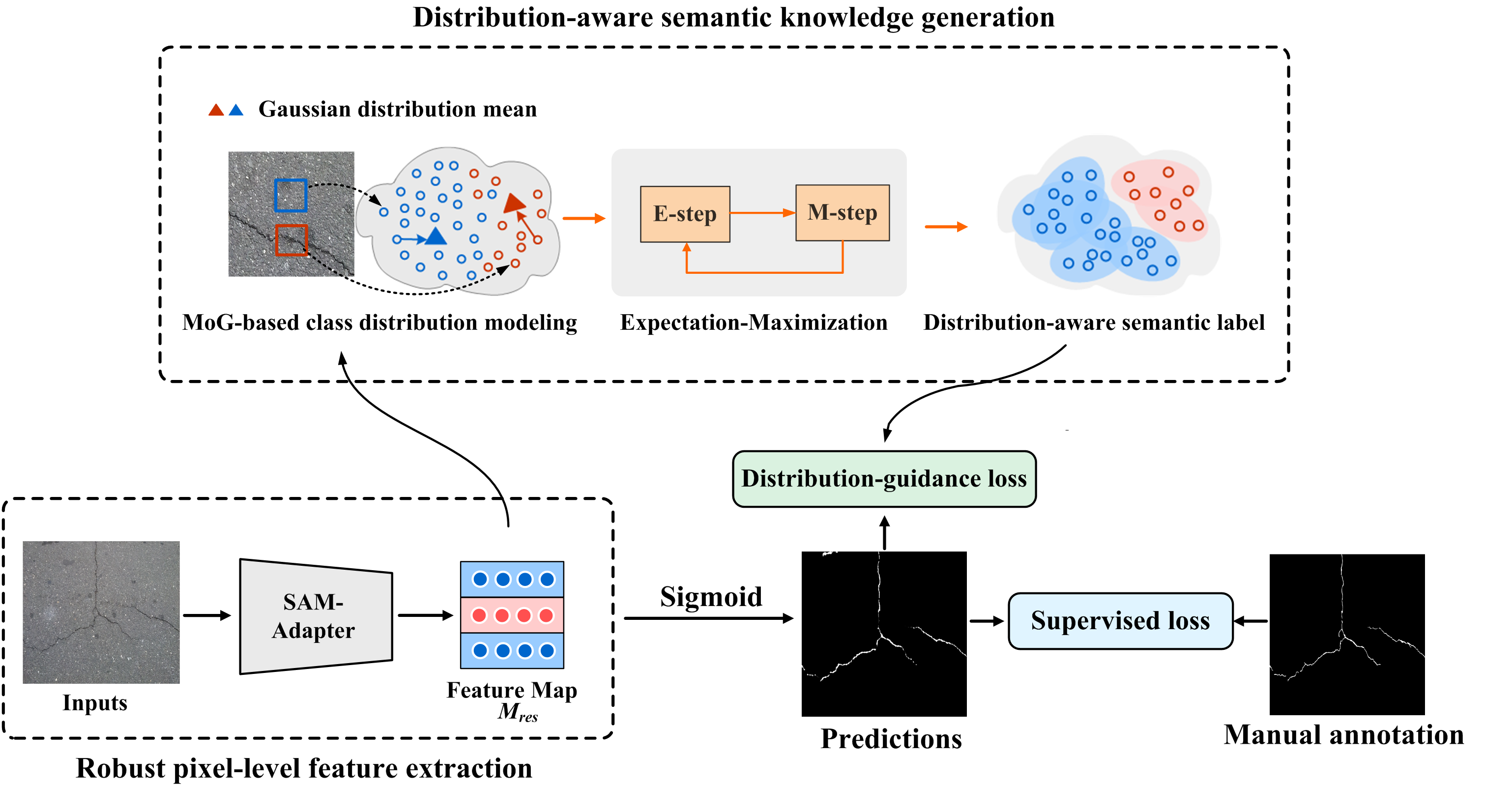}
	\end{minipage}
          \vspace{1.5mm}
 	\vspace{1.5mm}
    \caption{The proposed crack segmentation framework. 
    %Pixel level representations $M_{\text {res}}$ are extracted by SAM-Adapter with both versatile and task-specific knowledge. 
    %The domain-specfic distribution-aware semantic Knowledge Guidance Module absorbs $M_{\text {res}}$ as input and finds the optimal means and covariance for Gaussian Mixture Models in an unsupervised manner. 
    The network is jointly trained by both the supervised loss and the domain-specific distribution-guidance loss. }
    \label{Fig3}
\end{figure}

Fig. \ref{Fig3} outlines the proposed framework. It mainly includes the pixel-level semantic representation extractor module \ref{subsec:B} and the semantic knowledge guidance module \ref{subsec:C}.

\subsection{Preliminary: SAM-Adapter}
The SAM model \cite{ref9} is a powerful and versatile pre-trained large-scale image segmentation model that demonstrates excellent performance in numerous segmentation tasks. 
However, direct applying SAM to crack segmentation faces multiple challenges since the crack training data are not well represented in the training of SAM. 
To address this, SAM-Adapter \cite{ref10} can incorporate domain-specific information or visual cues into the segmentation network using a simple yet effective learnable module, Adapter. 
%SAM-Adapter is a parameter-efficient fine-tuning approach, it does not require fine-tuning the SAM network.
%The core concept of SAM-Adapter is to inject task-specific information into SAM using the learnable parameter Adapter. 
The task-specific information can be features extracted from task samples.
Specifically, the learnable parameter Adapter conveys information through the following formula \cite{ref10}:

\begin{equation}
\label{deqn_ex1}
P_{i}=\mathrm{MLP}_{up}\left(\mathrm{GELU}\left(\mathrm{M L P}_{tune}^{i}\left(F_{i}\right)\right)\right)~,
\end{equation}
where $\mathrm{M L P}_{tune}^{i}$ represents the $i^{th}$ linear layer that generates task-specific hints for each adapter, $\mathrm{M L P}_{u p}$ represents the upper projection layer used to adjust the feature dimension of the converter.
$F_{i}$ can be various forms, such as, the texture or frequency or combination. While $P_{i}$ is a hint generated by the Adapter to guide the model to task-specific adjustments in each layer of converters. $\mathrm{GELU}$ is the GELU activation function.
By integrating task-specific knowledge and the general knowledge, SAM-Adapter can significantly enhance its performance in downstream tasks. 

\subsection{Robust pixel-level feature extraction}
\label{subsec:B}
SAM possesses extensive general feature extraction capabilities and is capable of capturing high-level general semantic information of images. 
As shown in the bottom-part workflow of Figure \ref{Fig3}, crack images are input to SAM-Adapter to extract the pixel-level feature maps $M_{ {res }}$ in a discriminative learning manner.
Each data point of $M_{res}$ indicates the class it belongs to, which is a a pixel level semantic representation of the image sample.
Notably, the learning of $M_{res}$ absorbs evaluations (loss) from both supervised branch and the distribution-aware semantic knowledge guidance module, see \ref{subsec:D}.
Consequently, our crack SAM-Adapter integrates the following three information: the general knowledge from SAM, the  domain-specific knowledge of cracks by manual annotations, and the distribution guided semantic knowledge labels.
%Suppose the input image is $I$ and the output after the SAM-Adapter segmentation network is a pixel feature map $M_{ {res }}$.
% \begin{equation}
% M_{\text {res }}= Extractor(I)~,
% \end{equation}
% where $Extractor(.)$ represents the pixel-wise semantic representation extractor and is the SAM-Adapter semantic segmentation network. 

\subsection{Distribution-aware semantic knowledge generation}
\label{subsec:C}
As shown in upper Figure \ref{Fig3}, the unsupervised distribution-aware semantic knowledge generation module mainly includes the MoG-based class distribution modeling, the Expectation-Maximization (EM) algorithm, and the distribution-aware semantic knowledge labeling.

\textbf{MoG-based class distribution modeling.} 
To accurately mine the distributional properties of the crack and background, we select the Gaussian distribution to model the robust feature representations extracted by SAM-Adapter.
%In contrast to papers that  GMMSeg \cite{},
For each image sample, the pixel-level feature representation of every class is represented as a probabilistic distribution function, that is, Gaussian distribution. 
%to infer the distributional properties of cracks and backgrounds in the semantic space, thus generating semantic knowledge labels based on the distributional properties of the classes.
In our case, we have two classes, that is, crack and background, each of which subject to the following Gaussian distribution:

\begin{equation}
\label{equ:2}
p\left(\mathrm{\mathbf{x}} \mid \mathrm{c}_{k}^j \right) \sim \mathcal{N}\left(\mathbf{\mu}_{k}^j, \mathbf{\sigma}_{\mathrm{k}}^j \right)~,
\end{equation}
where $\mathbf{\mu}_{k}^j$ and $\mathbf{\sigma}_{\mathrm{k}}^j$ represent the mean vector and covariance matrix of features of class $k \in \{1,2\}$ in $j^{th}$ image sample, respectively.

To better reflect the complex distribution characteristics of cracks and background, we adopt Mixture of Gaussian distributions (MoG) instead of one single Gaussian for one class. 
Statistically, MoG can depict the semantic distribution of various crack types more accurately. 
Meanwhile, each Gaussian (equ. \ref{equ:2}) from one image keep the individual crack sample's characteristic.

Specifically, MoG for crack or background of all training data is defined as follows:
\begin{equation}
\label{equ:3}
p\left(\mathbf{x} \mid c_{k} \right) \sim \sum_{j=1}^{J} \pi_{j}\mathcal{N}\left(\mathbf{\mu}_{k}^{j}, \mathbf{\sigma}_{k}^{j}\right)~,
\end{equation}
where $\pi_{j}$ represents the weight of the $j^{th}$ Gaussian component for class $k$. $J$ is the total number of samples in training. 

\textbf{Expectation-Maximization (EM).} 
The EM algorithm is adopted to estimate the parameters of MoG for each $k$ shown in equ. \ref{equ:3}, including the mean vector, the covariance matrices, and the weights.
Since we know the observation $\mathbf{x}$ of each image, EM can find the parameters in probabilistic models with latent variables, which fits well with our task.
Note that, EM is an unsupervised optimization algorithm not disturbed directly by the noise labels in samples. 
EM calculates the expected values of the latent variables under the current parameter estimates and uses these expected values to update the model parameters to maximize the log-likelihood function in an iterative E-step and M-step.

In the E-step, the probability $\gamma_{i k}$ of each data point in the $j^{th}$ sample belonging to which Gaussian component of class $k$ is estimated based on the current parameters, as follows:

\begin{equation}
\gamma_{ik} = \frac{\pi_k^j \left|\mathbf{\sigma}_k^j\right|^{-1 / 2} \exp \left(-\frac{1}{2} D_k^j(\mathbf{x}_i)\right)}{\sum_{j=1}^{J} \pi_k^j \left|\mathbf{\sigma}_k^j\right|^{-1 / 2} \exp \left(-\frac{1}{2} D_k^j(\mathbf{x}_i)\right)}~,
\end{equation}
where $D_k^j(\mathbf{x}_i) = \left(\mathbf{x}_i - \mathbf{\mu}_k\right)^{T} \sigma_k^{-1} \left(\mathbf{x}_i - \mu_k\right)$ represents the Mahalanobis distance between data point $\mathbf{x}_{i}$ in the $j^{th}$ image sample and the corresponding Gaussian distribution mean. 
By this, correlation between data points are judged in the semantic representation space $M_{res}$, which helps the training to pull closer the points belongs to the same class.

In the M-step, we use $\gamma_{i k}$ to update the current model parameters:
\begin{equation}
\pi_{k}=\frac{1}{N} \sum_{i=1}^{N} \gamma_{i k}
\end{equation}
\begin{equation}
\mu_{k}=\frac{\sum_{i=1}^{N} \gamma_{i k} x_{i}}{\sum_{i=1}^{N} \gamma_{i k}}
\end{equation}
\begin{equation}
\sigma_k = \frac{1}{N_k} \sum_{i=1}^{N} \gamma_{ik} \left(x_i - \mu_k\right) \left(x_i - \mu_k\right)^{T}~,
\end{equation}
where $N$ is the total number of data points in the feature  space.

Iterative optimization process: we repeat the E-step and M-step until the model converges in the condition that the parameters to be estimated no longer change significantly.
Since the distribution-aware semantic knowledge guidance module only exits during training, our model has no inference speed increasing compared with SAM-Adapter.

\textbf{Distribution-aware semantic label.} 
Obtained the estimated distribution model of feature representations, we can generate distribution-aware semantic knowledge labels for image points.
%Obeys to the Bayesian rule, 
We estimate the probability of each feature representation belonging to a category in the pixel-level semantic representation map through the posterior probability. 
The posterior probability $p\left(c_{k} \mid \mathbf{x} \right)$ represents the likelihood that the semantic representation $\mathbf{x}$ belongs to the $k^{th}$ class (crack or background), shown as the following formula:
\begin{equation}
p\left(c_k \mid \mathbf{x}\right) = \frac{ \sum_{j=1}^{J} \pi_k^j \mathcal{N}\left(\mathbf{x} \mid \mu_k^j, \sigma_k^j\right)}{\sum_{{k'}=1}^{2} \sum_{j=1}^{J} \pi_{k'}^j \mathcal{N}\left(\mathbf{x} \mid \mu_{k'}^j, \sigma_{k'}^j\right)} ~.
\end{equation}

For each pixel, the distribution-aware semantic label $Label$ is determined by the maximum a posteriori probability as follows:
\begin{equation}
Label = \operatorname{argmax} \, p\left(c_k \mid x\right)
\end{equation}

%So far, we can model the distribution of the semantic representation, learn the joint probability distribution between semantic representations and category labels.
So far, robust distribution-aware semantic labels are generated to create high-quality semantic knowledge labels to guide the discriminative learning of crack segmentation with noisy labels.

\subsection{Loss Function}
\label{subsec:D}
\textbf{Distribution-guidance loss.} 
Normally, cracks only occupy small regions of the whole image, most image parts are backgrounds, which even have crack-like features in many cases, for example, asphalt and concrete roads. 
Since we have severe data imbalance, the Dice Loss is selected to calculate the distance between the prediction of SAM-Adapter and the distribution-aware semantic labels.
%As we mainly target under-labeled samples, Dice Loss is adopted for semantic knowledge guidance during model training. 
%, a loss function commonly used for handling class imbalance, is especially suitable for semantic segmentation. 
%It tends to explore the foreground (i.e., crack areas) and measures similarity by calculating the overlapping area (intersection) between the prediction and the true label, 
Distribution guidance loss $L_{Dg}$ is defined as follows:
\begin{equation}
L_{Dg}=1-\frac{2|X \cap Y|}{|X|+|Y|}~,
\end{equation}
where $X$ is the model prediction, that is, the probability distribution of the model output; $Y$ is the distribution-aware semantic label.
This makes the model focus on the small but important foreground crack areas during training and effectively mitigating class imbalance. 

\textbf{Supervised Loss.} In order to effectively alleviate the class imbalance problem and improve the stability of the model during training, we use the combination of binary cross-entropy loss function (BCE) and Dice Loss as the supervised loss. 
Binary cross-entropy calculates the loss for each pixel independently, which can provide reliable gradients for stable training. 
The BCE loss formula is as follows:
\begin{equation}
L_{Bce} =-\frac{1}{N} \sum_{i=1}^{N} q_{i} \log \left(p_{i}\right)+\left(1-q_{i}\right) \log \left(1-p_{i}\right)~,
\end{equation}
where $p_{i}$ and $q_{i}$ are the predicted probability value and the annotated label of the $i^{th}$ pixel, respectively. N is the total number of pixels in the image.
The total supervision loss $L_{Sup}$ is:
\begin{equation}
\label{equ:12}
L_{Sup} = {L_{Bce}}+\beta \cdot {L_{Dice}}~,
\end{equation}
where $\beta \in (0,1)$.

The total loss is:
\begin{equation}
\label{equ:13}
L_{total} = L_{Sup} + \lambda  \cdot L_{Dg} .
\end{equation}
%This combination is capable of providing global stability while attaching greater attention to the foreground regions, thereby enhancing the overall segmentation performance.

\section{EXPERIMENTS}
% This section initially presents the experimental Setting, the datasets utilised, the evaluation metrics, the model comparison outcomes, visual analysis, and ablation experiment analysis.
\subsection{Experimental Configuration}
To ensure the comparison fairness, we have the following uniform configuration for all the methods in the training stage. The size of the input image is 1024 x 1024. Random rotation and random flipping are conducted for data augmentation. All experiments use the AdamW optimizer and the same loss function. 
The initial learning rate is set to be 0.0003 and the cosine decay method is adopted to adjust the learning rate. 
The training period for each model was set to 90 epochs. 
$\beta$ in equ. \ref{equ:12} and $\lambda$ in equ. \ref{equ:13} are both set to be 0.3.
All experiments were implemented using PyTorch on an Intel Xeon Gold 6226R CPU and 1 NVIDIA Tesla A100 GPU.

\subsection{Datasets}
The CRACK500 dataset \cite{ref8} and the CFD dataset \cite{ref39} are the most commonly used public datasets for pixel-level segmentation of pavement cracks, which were collected by different camera configurations of different countries. 
The total captured images of the two datasets are 500 and 118, repectively.
To measure the model's crack recognization ability and cross-domain performance, \textit{we only use the training set of Crack500 for model training. }
For testing, the model is evaluated on the test set of Crack500 and the entire CFD dataset, respectively. 
Hence, we did not use any data from the CFD dataset for training.

%The CRACK500 dataset comprises 500 pavement crack images with 2000x1500 pixels' resolution, which were collected using a smart phone on the main campus of Temple University in the United States. Each image has undergone pixel-level binarization annotation, among which 250 images are for training, 50 for validation, and 200 for testing. It is currently the largest publicly available pavement crack dataset with pixel-level annotations. The road crack samples in the crack500 dataset include thick and thin cracks, but mainly thick cracks.

%The CFD dataset is a collection of images under the urban pavement conditions in Beijing, containing a total of 118 pavement crack pictures with the size 480x320 pixels. Each image was captured using an iPhone5 mobile phone and also underwent pixel-level binarization annotation. Note that, the road samples in the CFD dataset are mainly thin cracks.

\subsection{Evaluation Metrics}
% In the crack segmentation algorithm, the definitions of TP (true positive), FP (false positive), FN (false negative), and TN (true negative) for each category of pixels are very important. 
Due to the subjective differences in manual crack annotation, the discrimination error is generally allowed to be within 2 to 5 pixels at the edge of the area of the true crack pixels \cite{ref40}.
For evaluation, we use a small disc to dilate the true crack label, and the radius of the dilated disc is set to three pixels. 
This evaluation method prevents the penalty of the predicted crack pixels that are not fully aligned with the true crack and also absorbs the small positioning error at the edge of the crack \cite{ref24}.
To assess the performance of the model more comprehensively, we employ three prevalent and efficacious evaluation metrics for semantic segmentation, that is, F1 Score, Intersection over Union (IoU), and Dice Coefficient. 

% F1 Score is the harmonic mean of Precision and Recall, which integrates the advantages of both, and can effectively measure the accuracy and recall of the model in identifying cracks. It is defined as:

% \begin{equation}
% F 1=2 \times \frac{{ Precision } \times { Recall }}{{ Precision } + { Recall }}~.
% \end{equation}

% IOU is an important indicator to evaluate the overlap degree between the predicted results and the true results.
% %It is defined as the area of the intersection of the predicted region and the true region divided by the area of the union of them. 
% The formula is as follows:

% \begin{equation}
% I o U=\frac{T P}{T P+F N+F P}~,
% \end{equation}
% where TP indicates true positive, FN represents false negative, and FP is false positive.

% Dice coefficient, as defined in equ. \eqref{dice}, pays more attention to the ratio of the double area of the overlapping area to the total area, which is especially suitable for measuring the performance of small object segmentation tasks.
% \begin{equation}
% \label{dice}
% Dice =\frac{2 \times TP}{2 \times TP + FN +FP}~.
% \end{equation}

\subsection{Experimental results}
% Please add the following required packages to your document preamble:
\begin{table}
\renewcommand{\arraystretch}{1.5}
\setlength{\tabcolsep}{5pt}
% \normalsize
% \large
\caption{Comparison on the Crack500 test dataset and  CFD dataset(\%)}
\label{tabel1}
\centering
\begin{tabular}{c|ccc|ccc}
\hline
\multirow{2}{*}{Method} & \multicolumn{3}{c|}{Crack500} & \multicolumn{3}{c}{CFD} \\ \cline{2-7} 
                        &                               F1       & IoU      & Dice    & F1     & IoU    & Dice  \\ \hline
Unet \cite{ref16}                    & 58.47    & 42.64    & 57.05   & 11.31  & 7.11   & 11.08 \\
FCN \cite{ref15}                     & 45.24    & 30.29    & 43.77   & 3.18   & 2.06   & 3.46  \\
EMCAD \cite{ref41}                   &  63.57    & 46.13    & 61.33   & 31.69  & 22.2   & 33.47 \\ \hline
SegFormer \cite{ref21}               &   67.75    & 50.85    & 66.24   & 42.55  & 29.7   & 42.35 \\
HRViT \cite{ref20}                   &  61.76    & 45.75    & 60.93   & 16.42  & 10.87  & 16.53 \\
SAM-Adapter \cite{ref10}             &   75.23    & 57.91    & 72.66   & 66.55  & 48.92  & 63.88 \\ \hline
DeepCrack \cite{ref13}               &  71.54    & 55.28    & 70.28   & 46.2   & 32.6   & 45.81 \\
CrackFormer-II \cite{ref23}          &  72.12    & 55.41    & 70.48   & 58.78  & 42.34  & 57.2  \\ \hline
\textbf{Ours}                   &                              \textbf{76.66}    & \textbf{60.56}    & \textbf{74.74}   & \textbf{71.43}  & \textbf{53.37}  & \textbf{68.62} \\ \hline
\end{tabular}
\end{table}

% We compared the proposed approach with three principal methods: CNN-based methods including FCN \cite{ref18}, Unet \cite{ref17}, DeepCrack \cite{ref22}, and EMCAD \cite{ref48}; 
% Transformer-based methods including Segformer \cite{ref26}, HRvit \cite{ref25}, and CrackFormer-II \cite{ref31}; 
% The large-scale visual pre-training model adaptation-based, that is, SAM-Adapter \cite{ref13}; 
% Among them, CrackFormer-II obtains the highest performance in the domain of crack segmentation. 
% The outcomes of the model comparison experiments on the Crack500 test set and CFD dataset are presented in Table \ref{tabel1}. 
We compared the proposed approach with three main methods: CNN-based methods (FCN \cite{ref15}, Unet \cite{ref16}, DeepCrack \cite{ref13}, and EMCAD \cite{ref41}); Transformer-based methods (Segformer \cite{ref21}, HRViT \cite{ref20}, and CrackFormer-II \cite{ref23}); The large-scale visual pre-training model adaptation-based, that is, SAM-Adapter \cite{ref10}; Among them, CrackFormer-II achieved the best performance in crack segmentation. The results of the model comparison experiments on the Crack500 test set and CFD dataset are shown in Table \ref{tabel1}.

% The results on the Crack500 dataset: Table \ref{tabel1} shows that the method we proposed achieves the state-of-the-art performance and outperforms all CNN-based, Transformer-based, and including the second-best baseline large model adaptation method SAM-Adapter in terms of overall indicators (F1 value, IoU, and Dice) on the dataset. Specifically, compared with the (baseline) SAM-Adapter method, our method achieves performance increments of 1.43\%, 2.56\%, and 2.08\% in F1, IoU, and Dice coefficients, respectively, indicating that our method is effective. Using the knowledge of semantic representation to guide the model to suppress the interference of missed-labeled samples on the model can significantly improve the performance. Secondly, by comparing the baseline model SAM-Adapter with the previous CNN-based and Transformer-based methods, it can be found that integrating the general knowledge of SAM through adaptation with the domain knowledge of road surface crack segmentation can significantly improve the model's ability to identify cracks in this scenario and is superior to the current state-of-the-art method. Specifically, compared with the third-place current state-of-the-art method in this field, CrackFormer-II, SAM-Adapter improves by 3.11\%, 2.5\%, and 2.18\% in F1, IoU, and Dice coefficients, respectively, which also proves that the general large model has great potential for the road surface crack segmentation task. 

The results on the Crack500 test dataset in Table \ref{tabel1} show that our proposed method outperforms all CNN-based, Transformer-based and the leading large model adaptation method SAM-Adapter. Our method achieves state of the art performance on key metrics (F1 score, IoU, and Dice), showing improvements of 1.43\%, 2.56\%, and 2.08\% in F1, IoU, and Dice coefficients, respectively, over SAM-Adapter. The integration of semantic representation knowledge helps to mitigate the effects of mislabelled samples and significantly improves performance. In addition, SAM-Adapter's superior results over previous CNN and Transformer-based methods suggest that combining SAM's general knowledge with domain-specific insights improves crack identification capabilities.

The results of the CFD dataset in the UnSeen scenario are summarized in Table \ref{tabel1}, evaluating crack identification ability across scenarios.
All methods were trained on Crack500 and directly tested on the entire CFD dataset. Our proposed method significantly outperforms all existing state-of-the-art methods. Compared to SAM-Adapter, our approach improved F1, IoU, and Dice coefficients by 4.88, 4.45, and 4.74 respectively, highlighting its effectiveness in mitigating mislabeled samples and improving cross-domain crack detection.
Additionally, SAM-Adapter shows significant improvement in cross-scenario crack identification compared to previous CNN-based and transformer-based methods, highlighting the potential of large models in advancing deep learning applications for road surface crack segmentation.

% Additionally, the SAM-Adapter model, compared to previous CNN-based and Transformer-based methods, shows a substantial enhancement in cross-scenario crack identification, outperforming the third-place CrackFormer-II model by 7.77\%, 6.58\%, and 6.68\% in F1, IoU, and Dice coefficients, respectively. These findings underscore the significant potential of large models in advancing deep learning applications in road surface crack segmentation.
\begin{figure}[!t]
	\centering
 	\begin{minipage}{0.2\linewidth}
		\centering
		\includegraphics[width=1.8cm,height=1.6cm]{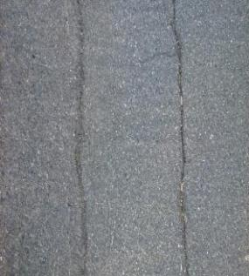}
	\end{minipage}
	\begin{minipage}{0.2\linewidth}
		\centering
		\includegraphics[width=1.8cm,height=1.6cm]{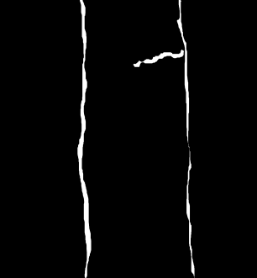}
	\end{minipage}
	\begin{minipage}{0.2\linewidth}
		\centering
		\includegraphics[width=1.8cm,height=1.6cm]{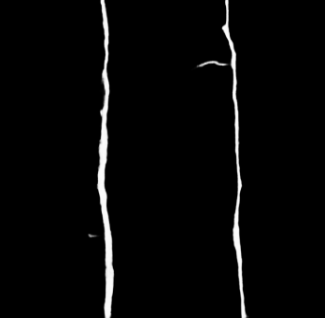}
	\end{minipage}
	\begin{minipage}{0.2\linewidth}
		\centering
		\includegraphics[width=1.8cm,height=1.6cm]{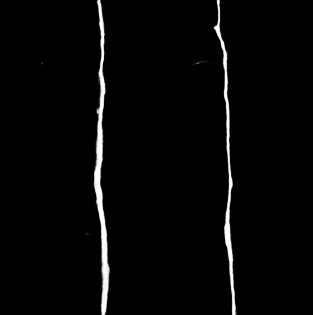}
	\end{minipage}
        \vspace{1mm}

        \centering
 	\begin{minipage}{0.2\linewidth}
		\centering
		\includegraphics[width=1.8cm,height=1.6cm]{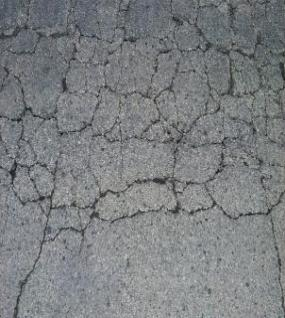}
	\end{minipage}
	\begin{minipage}{0.2\linewidth}
		\centering
		\includegraphics[width=1.8cm,height=1.6cm]{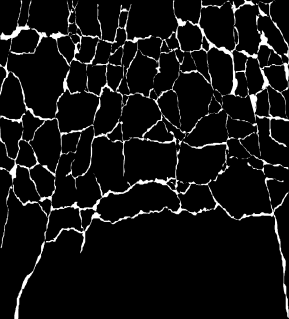}
	\end{minipage}
	\begin{minipage}{0.2\linewidth}
		\centering
		\includegraphics[width=1.8cm,height=1.6cm]{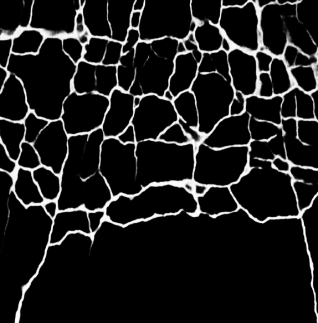}
	\end{minipage}
	\begin{minipage}{0.2\linewidth}
		\centering
		\includegraphics[width=1.8cm,height=1.6cm]{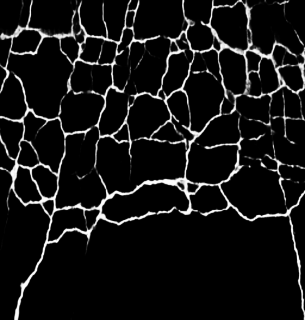}
	\end{minipage}
        \vspace{1mm}

        \centering
 	\begin{minipage}{0.2\linewidth}
		\centering
		\includegraphics[width=1.8cm,height=1.6cm]{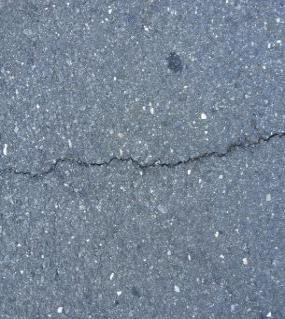}
	\end{minipage}
	\begin{minipage}{0.2\linewidth}
		\centering
		\includegraphics[width=1.8cm,height=1.6cm]{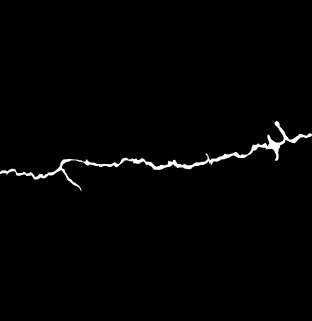}
	\end{minipage}
	\begin{minipage}{0.2\linewidth}
		\centering
		\includegraphics[width=1.8cm,height=1.6cm]{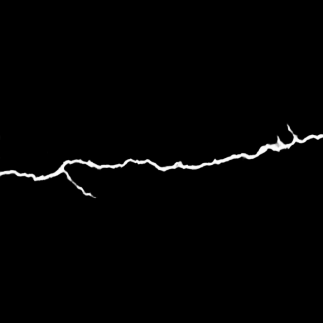}
	\end{minipage}
	\begin{minipage}{0.2\linewidth}
		\centering
		\includegraphics[width=1.8cm,height=1.6cm]{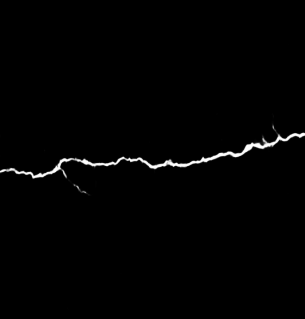}
	\end{minipage}
        \vspace{1mm}

	\begin{minipage}{0.2\linewidth}
		\centering
		\includegraphics[width=1.8cm,height=1.6cm]{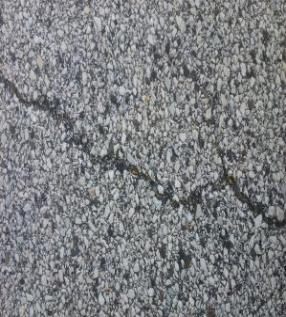}
		\centerline{\fontsize{7.5pt}{\baselineskip}\selectfont Original Image}
	\end{minipage}
	\begin{minipage}{0.2\linewidth}
		\centering
		\includegraphics[width=1.8cm,height=1.6cm]{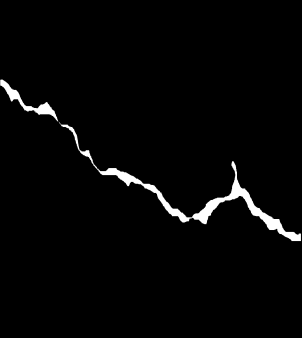}
  		\centerline{\fontsize{7.5pt}{\baselineskip}\selectfont Annotate labels}
	\end{minipage}
        \begin{minipage}{0.2\linewidth}
		\centering
		\includegraphics[width=1.8cm,height=1.6cm]{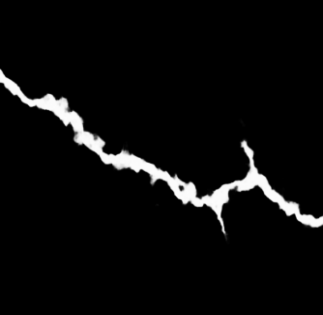}
            \centerline{\fontsize{7.5pt}{\baselineskip}\selectfont Ours}
	\end{minipage}
        \begin{minipage}{0.2\linewidth}
		\centering
		\includegraphics[width=1.8cm,height=1.6cm]{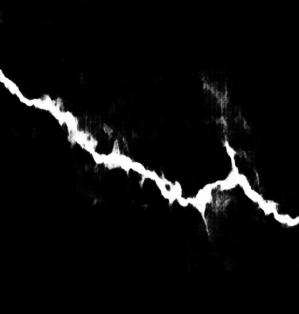}
            \centerline{\fontsize{7.5pt}{\baselineskip}\selectfont SAM-Adapter}
	\end{minipage}
 	% \vspace{1mm}
    \caption{Comparison on Crack500 test dataset}
    \label{Fig4}
\end{figure}
\vspace{1mm}

\begin{figure}[!h]
	\centering
 	\begin{minipage}{0.2\linewidth}
		\centering
		\includegraphics[width=1.8cm,height=1.6cm]{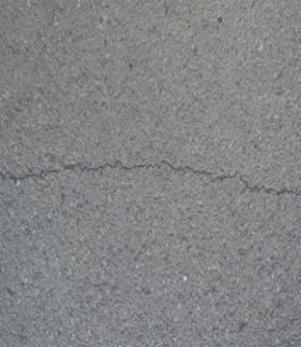}
	\end{minipage}
	\begin{minipage}{0.2\linewidth}
		\centering
		\includegraphics[width=1.8cm,height=1.6cm]{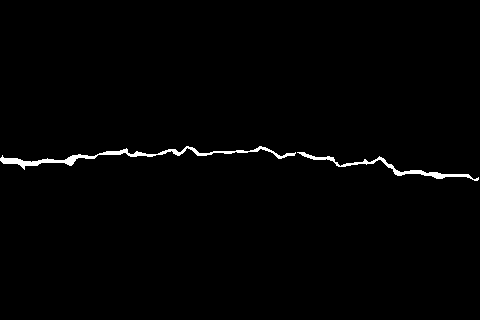}
	\end{minipage}
	\begin{minipage}{0.2\linewidth}
		\centering
		\includegraphics[width=1.8cm,height=1.6cm]{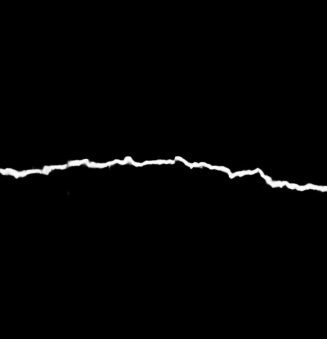}
	\end{minipage}
	\begin{minipage}{0.2\linewidth}
		\centering
		\includegraphics[width=1.8cm,height=1.6cm]{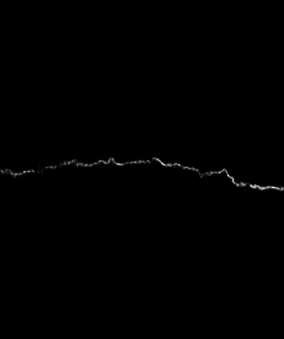}
	\end{minipage}
        \vspace{1mm}

        \centering
 	\begin{minipage}{0.2\linewidth}
		\centering
		\includegraphics[width=1.8cm,height=1.6cm]{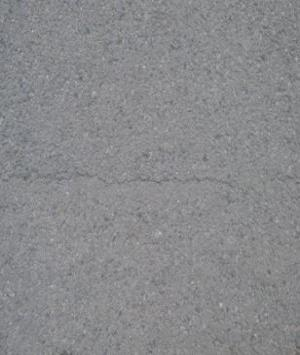}
	\end{minipage}
	\begin{minipage}{0.2\linewidth}
		\centering
		\includegraphics[width=1.8cm,height=1.6cm]{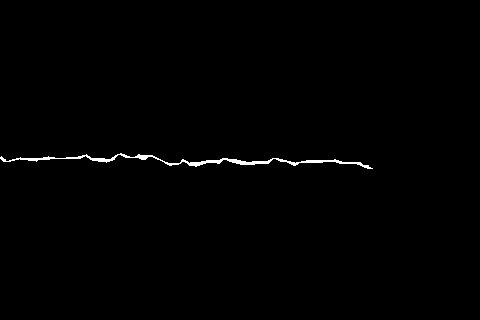}
	\end{minipage}
	\begin{minipage}{0.2\linewidth}
		\centering
		\includegraphics[width=1.8cm,height=1.6cm]{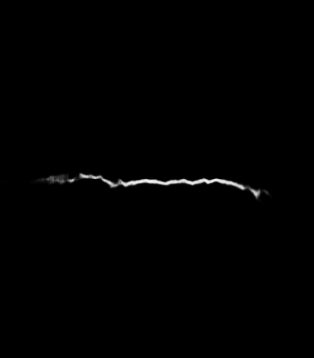}
	\end{minipage}
	\begin{minipage}{0.2\linewidth}
		\centering
		\includegraphics[width=1.8cm,height=1.6cm]{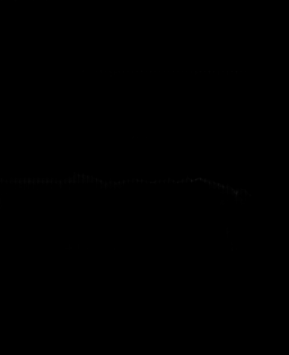}
	\end{minipage}
        \vspace{1mm}

        \centering
 	\begin{minipage}{0.2\linewidth}
		\centering
		\includegraphics[width=1.8cm,height=1.6cm]{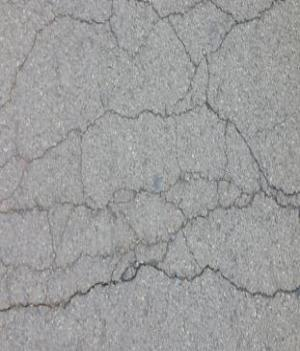}
	\end{minipage}
	\begin{minipage}{0.2\linewidth}
		\centering
		\includegraphics[width=1.8cm,height=1.6cm]{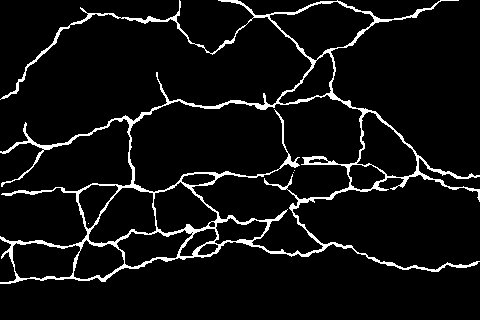}
	\end{minipage}
	\begin{minipage}{0.2\linewidth}
		\centering
		\includegraphics[width=1.8cm,height=1.6cm]{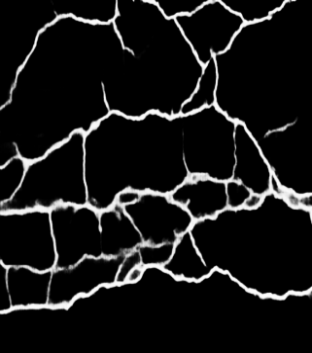}
	\end{minipage}
	\begin{minipage}{0.2\linewidth}
		\centering
		\includegraphics[width=1.8cm,height=1.6cm]{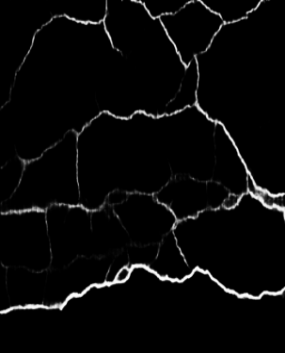}
	\end{minipage}
        \vspace{1mm}

         % %\qquad
	\begin{minipage}{0.2\linewidth}
		\centering
		\includegraphics[width=1.8cm,height=1.6cm]{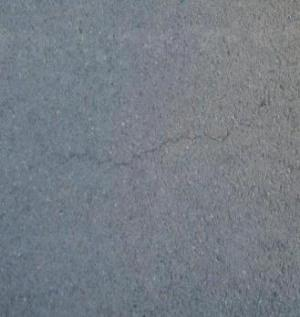}
		\centerline{\fontsize{7.5pt}{\baselineskip}\selectfont Original Image}
	\end{minipage}
	\begin{minipage}{0.2\linewidth}
		\centering
		\includegraphics[width=1.8cm,height=1.6cm]{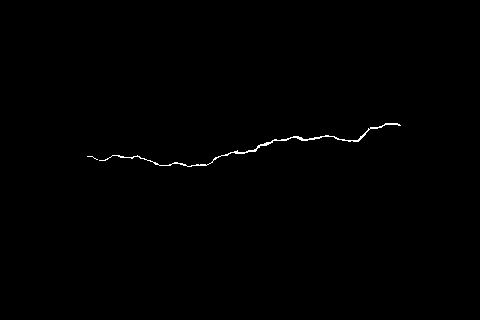}
  		\centerline{\fontsize{7.5pt}{\baselineskip}\selectfont Annotate labels}
	\end{minipage}
        \begin{minipage}{0.2\linewidth}
		\centering
		\includegraphics[width=1.8cm,height=1.6cm]{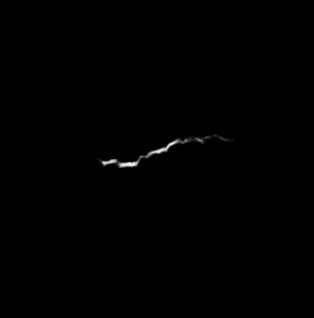}
            \centerline{\fontsize{7.5pt}{\baselineskip}\selectfont Ours}
	\end{minipage}
        \begin{minipage}{0.2\linewidth}
		\centering
		\includegraphics[width=1.8cm,height=1.6cm]{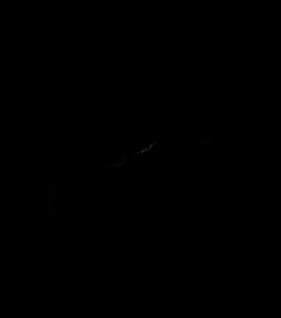}
            \centerline{\fontsize{7.5pt}{\baselineskip}\selectfont SAM-Adapter}
	\end{minipage}
         % \vspace{2mm}
 	% \vspace{1mm}
    \caption{Comparison of different methods' zero-shot ability on the CFD dataset.}
    \label{Fig5}
\end{figure}

\subsection{Visualization}
To visually present the proposed advantages, we visualize the prediction results on the CRACK500 and CFD datasets. As shown in Figure \ref{Fig4}, compared to the method in the second place, the method we proposed accurately identified some extremely fine forked cracks, such as the fine crack fork samples on the left side of the image (row 1) and (row 3). Additionally, in some areas with highly complex pavement backgrounds, such as the image samples in row 2 (large-area fine cracks) and row 4 (messy background colors), our method provided more accurate predictions and suppressed the interference of the complex background. This intuitively shows that in the training environment with real noisy labels, our method can effectively suppress the negative influence of missed-labeled data and perform more robustly in complex pavement backgrounds.

To intuitively demonstrate the identification ability of our proposed method for fine cracks in cross-domain, we visualized the results of the CFD dataset, as shown in Figure \ref{Fig5}. For some very fine cracks in cross-domain, our method is visually significantly superior to the second-place SAM-Adapter (baseline) method. Specifically, for example, in the second and fourth rows of the image, due to the interference of the missed-labeled samples in the training set of Crack500, the second-place SAM-Adapter method completely fails to identify the crack, while our method can identify the general outline of the crack. For the thin crack samples in the cross-domain, our method almost completely captures the details of the cracks, while SAM-Adapter misses many subtle details in the crack samples.
\subsection{Ablation Study}
% In order to evaluate the influence of each module on the metrics of the SAM model in the cross-dataset scenario, we use the model adapted from the Crack500 training set and conduct ablation experiment analysis on the CFD dataset. Specifically, we compare the original SAM model, the SAM-Adapter with Adapter addition, and the SAM-Adapter with semantic knowledge guidance module, and the experimental results are shown in Table \ref{tabel2}.
% (1) Adapter Module: We introduced the Adapter module in SAM to integrate the domain knowledge of crack segmentation with the general knowledge of the SAM large model. To prove the effectiveness of the Adapter method, we trained an original SAM model and a SAM-Adapter network with only the Adapter module. It can be seen from Table \ref{tabel2} that the model increased by 4.95\%, 4.49\%, and 4.07\% in the F1, IoU, and Dice coefficients, respectively, indicating that the Adapter module can improve the SAM model. 
% (2) Semantic Knowledge Guidance Module: We added a semantic knowledge guidance module based on SAM-Adapter to suppress the negative impact of noise labels on the model during training. To prove the effectiveness of the semantic knowledge guidance module, we trained a SAM-Adapter model and a model with the semantic knowledge guidance module. It can be seen from Table \ref{tabel2} that the model increased by 4.88\%, 4.45\%, and 4.74\% in the F1, IoU, and Dice coefficients, respectively. Therefore, adding the semantic knowledge guidance module can effectively suppress the negative impact of noise labels.

To assess each module's impact on SAM model metrics in a cross-dataset scenario, we conducted ablation experiments using the Crack500-trained model on the CFD dataset. We compared the original SAM model, SAM-Adapter, and SAM-Adapter with a semantic knowledge guidance module. Results in Table \ref{tabel2} show that the Adapter module alone improved F1, IoU, and Dice by 4.95\%, 4.49\%, and 4.07\%, respectively, indicating its effectiveness. Adding the semantic knowledge guidance module further enhanced these metrics by 4.88\%, 4.45\%, and 4.74\%, effectively mitigating noise label impact.

\begin{table}
\caption{Ablation study for key modules (\%)}%
\renewcommand{\arraystretch}{1.25}
% \normalsize
\label{tabel2}
\centering%
\begin{tabular}{ccccc}
\toprule
Method&F1&IoU&Dice \\
\midrule%
% \vspace{1.25mm}
SAM	&61.60&	44.43&	59.81\\
% \vspace{1.25mm}
+ Adpater	&66.55	&48.92	&63.88\\
+ MoG Module &71.43&	53.37	&68.62\\
\bottomrule%
\end{tabular}
\end{table}
\section{CONCLUSION}
% In this work, we re-examined the road surface crack segmentation task by considering the noise label issue of crack samples and designed a training framework based on general semantic knowledge guidance. This framework aims to enable the model to adaptively suppress its negative impact on training datasets with noise labels and improve the generalization and robustness of the model cross-domain.Technically speaking, we proposed a semantic representation modeling-based strategy to generate semantic knowledge labels, and encouraged the model to break away from overfitting to noise labels through joint supervision, so as to guide the model to adaptively suppress the negative impact of noise labels on the model during the training process. Evaluation on Crack500 and CFD datasets show that the semantic knowledge guidance module not only significantly improved the performance of SAM-Adapter but also achieved the state-of-the-art (SOTA) results, especially greatly enhancing the identification ability of SAM-Adapter for thin cracks in the cross-domain scenario. 

In this study, we re-examined the task of segmenting cracks from road images by explicitly addressing the issue of noise labels.
We developed a training framework based on distribution-aware general semantic knowledge guidance to improve the model's robustness and generalization ability. 
Our approach involved using semantic representation modeling to generate knowledge labels and implementing joint supervision to help the model avoid overfitting to noise labels during training. 
Evaluation on Crack500 and unseen CFD datasets demonstrated that our approach not only significantly improved the performance of SAM-Adapter in the downstream task but also achieved new state-of-the-art results, particularly enhancing its ability to identify thin cracks in cross-domain scenarios. 
We believe that our framework is beneficial for cracks analysis in other scenarios, such as, manufacturing quality control, pipeline inspections, and aircraft maintenance.

%\addtolength{\textheight}{-12cm}   % This command serves to balance the column lengths
                                  % on the last page of the document manually. It shortens
                                  % the textheight of the last page by a suitable amount.
                                  % This command does not take effect until the next page
                                  % so it should come on the page before the last. Make
                                  % sure that you do not shorten the textheight too much.

\bibliographystyle{ieeetr}
\bibliography{bibliography}
\end{document}